\documentclass[11pt]{article}

\usepackage[margin=1in]{geometry}
\usepackage[T1]{fontenc}
\usepackage[utf8]{inputenc}
\usepackage{amsmath}
\usepackage{booktabs}
\usepackage{graphicx}
\usepackage{float}
\usepackage{microtype}
\usepackage[round]{natbib}
\usepackage[hidelinks]{hyperref}

\title{Chain-of-Thought Entropy as a Reliability Signal:\\A Preregistered Reproduction}
\author{Theodore O. Cochran\\AI for Altruism\\\texttt{theo@ai4altruism.org}}
\date{}

\begin{document}
\maketitle

\begin{abstract}
This empirical study is an independent reproduction of the dissociation Zhao reported in 2026. The shape of a large language model's chain-of-thought entropy trajectory predicts whether the final answer is correct, while the magnitude of its total entropy drop does not. The dissociation merits reproduction because the magnitude half rests on a single 300-problem run with one model at one seed, while the shape half was reported at full scale on both benchmarks and on a second model family. Registered at OSF before any confirmatory run, the reproduction crosses the complete GSM8K and MATH-500 benchmark test sets with four open-weight models including one reasoning-distilled model of a kind the original did not test. The shape signal replicates. The magnitude signal divides by setting. On the anchor model the accuracy gap between monotone and non-monotone chains is $+9.6$ percentage points on GSM8K and $+27.5$ on MATH-500, while the rank correlation of the total entropy drop with correctness is $-0.018$ on GSM8K and $+0.414$ on MATH-500. On the reasoning-distilled model the binary form of the shape signal fires on about one chain in a hundred, too few to estimate the registered contrast, while the graded violation count remains predictive there. In an exploratory comparison the final-step entropy alone outperforms the binary shape flag in all eight model-by-benchmark cells by ROC area, and in six or seven by the risk-coverage area the original reports, depending on an integration range the original does not state. The study contributes a reproduction of the shape signal at full test-set scale under seven documented protocol differences, a map of the settings where the magnitude signal holds and fails, and measurements of four protocol dependencies the original does not report.
\end{abstract}

\section{Introduction}

Chain-of-thought prompting improves language-model accuracy on multi-step problems \citep{wei2022cot}, and it also makes failures harder to see, because an incorrect chain reads as fluently as a correct one. Methods that detect unreliable reasoning without sampling many full chains are therefore of practical interest, and the cheapest of them work from quantities that ordinary decoding already produces.

\citet{zhao2026entropy} proposes one such method. At each reasoning-step boundary of a single chain, a handful of short completions are sampled, the final answer is extracted from each, and the Shannon entropy of that answer distribution is recorded. The resulting per-step entropy trajectory is summarized in two ways: as a binary flag for whether entropy decreases at every step within a tolerance, which the paper calls shape, and as the scalar total drop from first step to last, which it calls magnitude. The reported result is a dissociation. Shape predicts correctness and magnitude does not, and the claim is that it is whether entropy falls at every step, not how far it falls in total, that carries the signal.

The claim is worth reproducing for three reasons. It is cheap enough to be adopted, at roughly 1,500 tokens per question against about 12,000 for forty-chain self-consistency \citep{wang2023selfconsistency}, so it is the kind of diagnostic that reaches production before it is independently checked. Its two halves rest on different amounts of evidence. The shape half is reported at $n=300$, at full scale on both benchmarks, and on a second model family, while the magnitude half is reported once, on 300 GSM8K problems with one model at one seed. And the original's own robustness sweep across three additional seeds returns accuracy gaps of $+8.3$, $+13.8$ and $+5.2$ percentage points against a headline of $+21.9$, which suggests the headline figure is less stable than a single number implies.

This study is a preregistered reproduction of that finding, extended to a reasoning-distilled model of a kind the original did not test. Four open-weight models were run over the complete GSM8K and MATH-500 test sets under a plan registered at OSF before any confirmatory run, with the analysis code frozen and public. The registered hypotheses cover the shape effect, the magnitude null, the graded violation-count signal, the behavior of the method on a reasoning-distilled model, and a learned false-positive-reduction filter.

Three contributions are claimed. The first is the reproduction itself. The shape signal is confirmed at full test-set scale across three standard instruct models and two benchmarks, under a protocol that differs from the original in seven specified ways, which makes the confirmation a statement about robustness rather than about pipeline agreement. The second is a division of the dissociation. Its violation half replicates in every cell, while its scalar-coherence half reproduces in the one cell matching the conditions the original measured it under and does not meet the registered criterion in six conditions for which the original reports no coherence figure. The third is a set of measurements of the protocol's dependencies. The reference-chain token budget, the segmentation rule, the step cap and the outcome definition each change what is measured, by amounts reported here and not reported in the original.

No independent reproduction of the finding was found. A search on 2026-09-05 of the Semantic Scholar citation graph, OpenAlex, arXiv metadata and the open web returned six papers citing \citet{zhao2026entropy}, none of which reproduces it. They carry entropy dynamics into inference routing \citep{xia2026dynamical}, into token-level failure signatures \citep{thoria2026failure}, and into the structure of chains on satisfiability tasks \citep{sourav2026telltale}. No replication report, reproduction code or registered replication appeared elsewhere, and the one open-web result that reads as a replication is Zhao's own cross-model check on Mistral. What this study claims, therefore, is that no reproduction was found on that date by those means.

Two neighbors bear on how the results below should be read. \citet{zhu2026edis} propose a trajectory-level instability score over token-level entropy, six weeks before Zhao's preprint and on the same underlying intuition, that the temporal evolution of entropy carries more than its aggregate does. Zhao's paper does not cite that work, so the shape-over-magnitude framing arrives without its nearest precedent in view. \citet{song2026prefixsafe} report, on the same two benchmarks used here, that probability quality and ranking quality separate, with ranking gains requiring structure-aware evidence that strong prefix-safe baselines have not already absorbed. That is the distinction Section~\ref{sec:scalar} meets from the other side, where a ranking statistic and the original's risk-coverage statistic disagree about whether shape beats a scalar.

\section{The Original Finding}
\label{sec:original}

Zhao's protocol generates one chain-of-thought chain per problem at temperature 0.1 with a 512-token cap, segments it into steps by matching the regular expression \texttt{Step [0-9]+:} and falling back to double-newline or sentence splitting when no marker is found, and then, at each cumulative step prefix, samples $m=5$ continuations at temperature 0.7 with a 150-token cap. The final answer is extracted from each continuation, and the Shannon entropy of the empirical answer distribution at prefix k gives $H_k$. A chain is $\epsilon$-monotone when $H_{k+1}$ is at most $H_k$ plus $\epsilon$ at every k, with $\epsilon$ fixed at 0.01 nats, and the scalar coherence is $C = H_0 - H_N$.

The reported results fall into two groups, and the distinction matters for what a reproduction can compare against. On a 300-problem GSM8K subset with Qwen2.5-7B-Instruct, monotone chains reach 68.8\% accuracy against 46.8\% for non-monotone chains, a gap of $+21.9$ percentage points with an odds ratio of 2.50 and a Fisher exact p of 0.0005, while scalar coherence has Spearman $\rho$ of $-0.059$ with correctness at p=0.31. That pilot is the only place the magnitude null is measured. At full scale, monotone chains reach 93.2\% against 81.7\% on GSM8K at $n=1319$, a gap of $+11.5$ points at monotone coverage 68.1\%, and 63.7\% against 30.4\% on MATH-500 at $n=500$, a gap of $+33.3$ points at coverage 27.0\%, with violation-count Spearman $\rho$ of $-0.198$ and $-0.381$ respectively. A cross-model check on Mistral-7B-Instruct-v0.3 at $n=300$ gives 72.3\% against 37.6\%, an odds ratio of 4.33.

Because the full-test-set figures are the ones this study's design matches, they are the comparison used throughout Section 4. Reporting the $+21.9$ point pilot gap as the target would overstate the distance between the two studies, since the original's own three-seed sweep on that subset averages $+9.1$ points. Two further results are relevant later. An $\epsilon$ ablation finds the monotone rate identical at 0.737 for every $\epsilon$ at or below 0.10, which the paper attributes to entropy jumps in non-monotone chains tending to exceed 0.20 nats. And a prefix analysis finds that using only the first two entropy transitions recovers $+16.7$ of the $+21.9$ point gap, or 76\%, at 60\% of the trajectory cost.

\section{Method}
\label{sec:method}

\subsection{Protocol}

The harness implements the protocol described above. For each problem it generates one reference chain, segments it into cumulative prefixes, samples $m=5$ continuations at temperature 0.7 with a 150-token cap at each prefix, extracts a final answer from each continuation, and computes the Shannon entropy of the resulting answer distribution in natural log units. From the trajectory it derives the binary $\epsilon$-monotonicity flag at $\epsilon=0.01$, the violation count, the coherence magnitude $C$, and a final-answer confidence taken as the self-consistency of the answer distribution at the last prefix. Correctness is the majority final answer at the last prefix compared against the normalized gold answer.

Answer extraction is by dataset. GSM8K answers are the final number in the completion, normalized by stripping currency symbols and thousands separators and dropping a trailing point-zero. MATH-500 answers are the last boxed expression under standard MATH normalization, falling back to the final number when no boxed expression is present. Entropy is computed in nats; the base rescales entropy uniformly and cannot affect a monotonicity test or a rank correlation, both of which are invariant to it.

\subsection{Model Panel and Datasets}
\label{sec:panel}

Four open-weight models were run at pinned Hugging Face revisions, all at bfloat16 through vLLM 0.28.0 on a single NVIDIA A40. Precision was held constant across the panel deliberately, so that the comparison between the reasoning-distilled model and the standard ones is not confounded by quantization.

\begin{table}[htbp]
\centering
\begin{tabular}{lll}
\toprule
Model & Revision & Role \\
\midrule
Qwen2.5-7B-Instruct & \texttt{a09a3545} & anchor, matching the original \\
Mistral-7B-Instruct-v0.3 & \texttt{c170c708} & matching the original's cross-model check \\
Llama-3.1-8B-Instruct & \texttt{0e9e39f2} & independent standard family \\
DeepSeek-R1-Distill-Qwen-7B & \texttt{916b56a4} & reasoning-distilled stress test \\
\bottomrule
\end{tabular}
\end{table}

The model families are described by \citet{qwen2024}, \citet{jiang2023mistral}, \citet{grattafiori2024llama3} and \citet{deepseek2025r1} respectively, with one caveat. The Mistral report covers the base model and its v0.1 instruct release, not the v0.3 instruct weights run here, which have no accompanying paper.

Both datasets were used complete: the GSM8K test split of \citet{cobbe2021gsm8k} at $n=1319$ and MATH-500 at $n=500$, the latter being the 500-problem subset of the MATH benchmark of \citet{hendrycks2021math} that \citet{lightman2023verify} drew and distributed, giving eight model-by-dataset cells and 7,276 intended records. The sample is a census of both benchmarks rather than a draw from them, which removes any item-sampling decision and makes the comparison with the original exact on that axis. It does not remove statistical uncertainty. Every interval below is a problem-level bootstrap, which quantifies between-item uncertainty conditional on the single realized decoding run, and says nothing about how much a rerun would move. No power analysis is offered, and effect estimates are reported with intervals so that the resolution actually achieved is visible.

\subsection{Differences From the Original Protocol}
\label{sec:differences}

\begin{table}[htbp]
\centering
\small
\begin{tabular}{p{0.20\textwidth}p{0.36\textwidth}p{0.36\textwidth}}
\toprule
 & \citet{zhao2026entropy} & This study \\
\midrule
Reference-chain temperature & 0.1 & 0.7 \\
Reference-chain token budget & 512 & 600 \\
Segmentation, primary rule & \texttt{Step [0-9]+:} regex with newline or sentence fallback & blank line \\
Step cap & none reported, mean 4.9 steps & 8 units \\
Continuations & $m=5$, $\tau=0.7$, 150 tokens & identical \\
Monotonicity tolerance & $\epsilon=0.01$ & identical \\
Decoding stack & HuggingFace transformers \texttt{model.generate} & vLLM 0.28.0 \\
Answer extraction, MATH & regex for integers and decimals at the end of a completion; no boxed path documented & last \verb|\boxed{...}| under MATH normalization, falling back to the final number \\
Correctness target & not stated; the self-judgment baseline is given ``its own final answer'', implying the chain's & majority answer of the last prefix's $m=5$ samples \\
\bottomrule
\end{tabular}
\end{table}

Seven parameters differ from Zhao's specification, and one further difference cannot be resolved at all. Each is stated because a reproduction that differs silently cannot be interpreted either way. The last two rows were identified after the panel ran and are definitional rather than implementation-level.

The reference-chain temperature difference is the consequential one, and it arises from an implementation decision rather than a design choice. The harness generates the reference chain with the same sampling temperature it uses for continuations. The concern was that a chain sampled at 0.7 might wander more than one sampled at 0.1, making it longer on average, less likely to converge, and more likely to reach the token cap. Section~\ref{sec:limitations} tests the length and truncation routes directly and finds no evidence for either, while the convergence route stays open. That sits directly upstream of the magnitude measure, and it is treated as a limitation rather than as a neutral difference.

The last two rows are definitions rather than parameters. Zhao's extraction is a regex over integers and decimals at the end of a completion, and he documents no boxed-expression path anywhere, including for his MATH-500 runs, while this study extracts the last boxed expression under MATH normalization. That plausibly bears on the MATH-500 accuracy gap, 50.5\% here against his 39.4\%, and it is not resolvable from the text. The correctness target is the more serious of the two. This study scores the majority answer of the last prefix's five samples, which is the same five the final entropy is computed from, while the original's phrasing in its self-judgment baseline implies it scores the reference chain's own answer. Neither point can be settled against an implementation, because no code accompanies the original paper, and Section~\ref{sec:scalar} measures what the coupling does to the results that depend on it.

The segmentation rule and the step cap are this study's own choices, fixed in advance rather than selected after seeing outcomes. Blank-line segmentation was registered as primary on the reasoning that the system prompt asks the model to put each reasoning step on its own line, and that a marker regex would fail differently on a model class that emits no markers at all. The first half of that reasoning does not survive Section~\ref{sec:segmentation}, so the rule stands as the registered choice but should not be read as a faithful reading of the prompt. The cap of 8 units was chosen to bound cost, since a chain with 40 units would require 41 prefixes and roughly 205 continuations.

The system prompt used throughout is:

\begin{quote}
\ttfamily
Solve the problem step by step. Put each reasoning step on its own line. End with the final answer.
\end{quote}

\noindent Zhao describes his own only as a chain-of-thought system prompt and does not give the text, so prompt equivalence cannot be established.

\subsection{Preregistration and Inference Criteria}
\label{sec:prereg}

The analysis plan was registered at OSF as registration \texttt{8w2q3} on 2026-08-29, from \texttt{docs/\allowbreak preregistration.md} at commit \texttt{a2edb8e9}, permanently reachable in the public repository as tag \texttt{prereg-v1.0}. Registration preceded every confirmatory run. The five registered hypotheses and their success criteria are:

\begin{itemize}
\item H1, replication, directional. Monotone chains are more accurate than non-monotone chains on both GSM8K and MATH-500 for the anchor model. Criterion: the accuracy-gap confidence interval excludes zero in the positive direction on both datasets.
\item H2, magnitude null, directional-null. Coherence magnitude is at most negligibly associated with correctness. Criterion: absolute Spearman $\rho$ below 0.10, and substantially smaller than the shape effect.
\item H3, graded signal, directional. Accuracy falls as the violation count rises. Criterion: $\rho$ below zero with a confidence interval excluding zero.
\item H4, reasoning-distilled behavior, estimation, explicitly two-sided. The accuracy gap for the distilled model and the difference between that gap and the mean gap of the three standard models, each with a confidence interval. Registered two-sided so that neither direction could become a post-hoc story.
\item H5, false-positive reduction, directional. A logistic filter over violation count and final-answer confidence, fit on a stratified 60\% training split, achieves a lower false-positive rate than the bare monotonicity baseline at matched coverage on the held-out 40\% test split. Criterion: the reduction interval excludes zero favorably.
\end{itemize}

The primary inferential object throughout is a 95\% percentile bootstrap confidence interval over 1000 resamples at the problem level, with p-values secondary. One exclusion rule was fixed in advance and implemented before registration. A problem is excluded when it yields fewer than two parseable answer-bearing steps, because a trajectory of fewer than two points admits no monotonicity determination. No other exclusion is permitted, and no imputation is performed anywhere.

The registration scopes those hypotheses more narrowly than a per-cell reading of Section 4 suggests, and the difference matters for what may be called confirmatory. Section 4 of the registration defines H1 on the anchor model only, and Section~\ref{sec:limitations} designates H1 and H4 as primary and lists H2, H3, H5 and the replication of H1 to H3 on the two additional standard models as secondary confirmatory. So of the eight cells, two are H1 as registered, four are registered secondary replication, and two belong to H4. Two features of that plan are worth stating rather than smoothing over. H2 is a null and is not in the Holm family, so a correction cannot help it in either direction. And the registration makes confidence intervals primary with p-values secondary while specifying a p-value procedure for multiplicity, which is resolved here by applying Holm to the $p$-value-bearing components of H1 and H3 and evaluating H5 separately against its registered interval criterion.

Holm-Bonferroni was registered and was not implemented in the as-run analysis code. It is applied here over the family the registration names, which is H1 on both anchor datasets, H3, and H5, counting H3 as one test per dataset. All four p-value-bearing tests clear their thresholds, the closest by roughly three orders of magnitude, and H5 meets its interval criterion on both anchor cells. No registered verdict changes under the correction, which is why the omission is reported as a deviation rather than as a defect. Appendix~\ref{app:holm} gives the registered scope, the arithmetic, and an earlier revision's sixteen-test panel-wide family, which is not the registered one.

\section{Results}
\label{sec:results}

Nine cells were run in total: the eight registered confirmatory cells and one exploratory cell described in Section~\ref{sec:budget}. The registered panel produced 7,275 of 7,276 intended records, of which 7,013 are usable after 262 exclusions under the fewer-than-two-steps rule. The exploratory cell adds 500 records and is excluded from every tally below, since it is outside the registration and cannot support or refute a registered hypothesis.

The table below gives the shape contrast for every cell. H1 as registered covers only the two anchor rows, the Mistral and Llama-3.1 rows are its registered secondary replication, and the distilled rows belong to H4, so the last column records whether each interval excludes zero and is not a hypothesis verdict.

\begin{table}[htbp]
\centering
\small
\setlength{\tabcolsep}{3.5pt}
\begin{tabular}{lrrrrrrlrl}
\toprule
cell & $n$ & excl. & acc & med. pts & cov. & gap (pp) & 95\% CI & OR & CI excl.\ 0 \\
\midrule
qwen7b-gsm8k & 1296 & 1.7\% & 0.846 & 5 & 0.563 & $+9.62$ & [$+5.58$, $+13.78$] & 2.08 & yes \\
qwen7b-math500 & 499 & 0.2\% & 0.505 & 7 & 0.146 & $+27.49$ & [$+16.56$, $+38.46$] & 3.27 & yes \\
mistral7b-gsm8k & 1134 & 14.0\% & 0.466 & 3 & 0.693 & $+20.74$ & [$+14.97$, $+26.70$] & 2.37 & yes \\
mistral7b-math500 & 457 & 8.6\% & 0.085 & 7 & 0.302 & $+7.50$ & [$+1.09$, $+14.05$] & 2.39 & yes \\
llama31-8b-gsm8k & 1316 & 0.2\% & 0.793 & 5 & 0.511 & $+21.86$ & [$+17.76$, $+26.42$] & 4.22 & yes \\
llama31-8b-math500 & 492 & 1.6\% & 0.309 & 8 & 0.157 & $+21.88$ & [$+9.17$, $+32.96$] & 2.57 & yes \\
r1-distill-7b-gsm8k & 1319 & 0.0\% & 0.340 & 8 & 0.013 & $+48.94$ & [$+28.67$, $+66.77$] & 9.30 & yes \\
r1-distill-7b-math500 & 500 & 0.0\% & 0.234 & 9 & 0.006 & $+9.99$ & [$-26.05$, $+77.96$] & 1.64 & no \\
\bottomrule
\end{tabular}
\end{table}

\subsection{The Shape Signal}
\label{sec:shape}

H1 as registered is supported. On the anchor model the accuracy gap is $+9.62$ points on GSM8K, at an odds ratio of 2.08 and a Fisher exact $p$ of 1.6e-06, and $+27.49$ points on MATH-500, at an odds ratio of 3.27 and $p=9.2$e-06. The criterion requires the interval to exclude zero positively on both datasets, and it does, before and after the Holm correction of Section~\ref{sec:prereg}.

The registered secondary replication holds as well. On the two additional standard models the same contrast is positive with an interval excluding zero in all four cells. The two distilled cells belong to H4 rather than to H1, and one of them carries the panel's only gap interval that includes zero. The result is therefore supported where registered, replicated in all four secondary cells, and not significant once, in a cell whose registered role is estimation.

Agreement with the original is close when it is set against the original's full-test-set figures rather than its 300-problem pilot. Zhao reports $+11.5$ points on full GSM8K and $+33.3$ on MATH-500, against $+9.62$ and $+27.49$ here on the same benchmarks with the same anchor model. Monotone coverage is lower here on both, at 0.563 against 0.681 and 0.146 against 0.270, which is what a segmentation rule producing different step counts would do. Overall accuracy differs in opposite directions, at 84.6\% here against roughly 89.5\% on GSM8K and 50.5\% against 39.4\% on MATH-500, so the two studies are not measuring identical model behavior even on the shared anchor.

That interval should be read as a failure of power rather than as evidence of absence. On r1-distill-7b-math500 the gap interval is 104 points wide, at [$-26.05$, $+77.96$], because the monotonicity flag fires on 0.6\% of chains, about three of 500. Section~\ref{sec:distilled} treats the collapse of coverage on that model as the finding.

\subsection{The Magnitude Null}
\label{sec:magnitude}

H2 is the registered directional-null claim. Across the full eight-cell panel its criterion is met in two cells and not met in six. Within the six standard-model cells that make up H2's registered scope, it is met in two and fails in four, and the two distilled cells are reported descriptively. Both cells that meet it are GSM8K cells, the anchor at $\rho$ $-0.018$ and Mistral at $+0.030$, each with an interval inside the criterion band of absolute $\rho$ below 0.10. Everywhere else the criterion is not met, and in four cells by a wide margin, reaching $+0.414$ on the anchor's MATH-500 cell and $+0.459$ on the distilled GSM8K cell.

The count of two in eight is misleading on its own, because the original speaks to only one of those conditions. Zhao reports a scalar-coherence figure exactly once, at $\rho$ $-0.059$ on the 300-problem GSM8K pilot with the anchor model, and none on full GSM8K, on MATH-500, or for any other model. The one cell of this panel that matches those conditions is qwen7b-gsm8k, and it reproduces his null at $-0.018$ with an interval of [$-0.074$, $+0.038$]. The six cells that miss the criterion are conditions in which the original publishes no coherence result to contradict. The coherence null therefore reproduces where it was measured and does not generalize beyond it.

\begin{table}[H]
\centering
\small
\begin{tabular}{lrll}
\toprule
cell & magnitude $\rho$ & 95\% CI & $|\rho| < 0.10$ \\
\midrule
qwen7b-gsm8k & $-0.018$ & [$-0.074$, $+0.038$] & met \\
qwen7b-math500 & $+0.414$ & [$+0.342$, $+0.489$] & not met \\
mistral7b-gsm8k & $+0.030$ & [$-0.031$, $+0.089$] & met \\
mistral7b-math500 & $+0.163$ & [$+0.086$, $+0.236$] & not met \\
llama31-8b-gsm8k & $+0.126$ & [$+0.071$, $+0.184$] & not met \\
llama31-8b-math500 & $+0.370$ & [$+0.297$, $+0.442$] & not met \\
r1-distill-7b-gsm8k & $+0.459$ & [$+0.409$, $+0.506$] & not met \\
r1-distill-7b-math500 & $+0.379$ & [$+0.297$, $+0.456$] & not met \\
\bottomrule
\end{tabular}
\end{table}

The anchor's MATH-500 cell is the sharpest failure, on the same model the original used and on a benchmark it ran at full scale for the shape signal without reporting coherence. Magnitude is not merely weakly predictive there. It is strongly predictive, with an interval of [$+0.342$, $+0.489$] nowhere near the criterion, and the shape effect on that cell is larger than on GSM8K, so the two measures are not competing for a fixed amount of signal.

Ordered by magnitude $\rho$, the six cells that miss the criterion line up with trajectory length better than with benchmark or accuracy. Mistral on GSM8K is the instructive case, since it has the lowest GSM8K accuracy of the three standard models at 46.6\% and yet holds H2, and it has the shortest trajectories at a median of 3 points. That is an observation about the ordering rather than a mechanism, and it was confounded with truncation until the exploratory cell and the budget scans separated the two. Chain length is already documented as a confounder in the original, which reports a Spearman $\rho$ of $-0.37$ between chain length and the monotonicity flag.

Whether the failures belong to the signal or to this study's instrument is addressed in Sections~\ref{sec:budget}, \ref{sec:segmentation} and \ref{sec:scalar}, which argue against the reference-token budget as the explanation, leave an endpoint asymmetry open, and report a third dependency in the correctness label itself. The other half of the original's dissociation, which concerns the magnitude of individual entropy violations rather than the total drop, is reproduced in Section~\ref{sec:scalar} and holds in every cell.

\subsection{The Graded Violation Count}

H3 is supported in all eight cells. Violation-count Spearman $\rho$ runs from $-0.108$ to $-0.338$, and every interval lies entirely below zero. On the anchor, $\rho$ is $-0.169$ on GSM8K against the original's $-0.198$, and $-0.338$ on MATH-500 against its $-0.381$, so the graded form reproduces about as closely as the binary one.

The bucket pattern on the anchor is monotone on both datasets. On GSM8K accuracy falls 88.8\%, 84.2\%, 69.9\% and 47.1\% across zero, one, two, and three or more violations, on 730, 419, 113 and 34 problems, and on MATH-500 it falls from 74.0\% to 29.4\% across the same buckets. The ordering holds in every other cell but one, where Mistral's top GSM8K bucket at 22.7\% sits above its two-violation bucket at 18.1\% on 44 problems, a difference well inside sampling noise at that count.

This is the quietest result in the panel and the most useful, since the graded count survives everywhere the binary flag becomes rare, including the distilled cells where binary coverage collapses to about 1\%.

\subsection{Reasoning-Distilled Models}
\label{sec:distilled}

H4 carried the study's registered novelty, and its estimand turns out not to be estimable with useful precision on the model it was written for. On DeepSeek-R1-Distill-Qwen-7B the monotonicity flag fires on 1.3\% of GSM8K chains, 17 of 1319, and on 0.6\% of MATH-500 chains, 3 of 500, against the anchor's 56.3\% on GSM8K. The registered contrast between the distilled gap and the mean gap of the three standard models computes to $+31.6$ points on GSM8K and $-9.0$ points on MATH-500, opposite in sign and resting on 17 and 3 monotone chains. Reporting either as an effect estimate would be false precision, and the registration made H4 two-sided estimation so that neither number could become a story after the fact.

The binary monotonicity criterion does not transfer usefully to this model, while the graded form of the same signal does. Violation count predicts correctness in both distilled cells, as it does in all eight, so what collapses is the binary flag as a selective predictor rather than the trajectory-shape signal itself. Two structural features plausibly contribute to the collapse. The distilled model emits long free-form chains, so its trajectories run at or near the maximum reachable length, from 7 to 9 points on GSM8K and 4 to 9 on MATH-500 under a cap of 8 units, and a longer trajectory has more opportunities to violate. Its chains also do not fit the token budget, which Section~\ref{sec:budget} quantifies. One model is not a class, and every distilled measurement here rests on DeepSeek-R1-Distill-Qwen-7B alone, so whether reasoning-distilled models in general behave this way is a hypothesis this study raises and does not test.

The 34.0\% accuracy on GSM8K should not be read as a property of the model. It is far below what DeepSeek-R1-Distill-Qwen-7B achieves on grade-school arithmetic, and the explanation is that the 600-token reference budget cuts off a large share of its chains before they reach an answer. That number measures the instrument.

\subsection{The Learned Triage Filter}
\label{sec:triage}

H5 was registered as an improvement at matched coverage, and the as-run analysis did not match coverage, so its verdict had to be recovered rather than read off. The as-run code thresholds on a quantile of predicted P(correct) and accepts everything above it, and because \texttt{final\_confidence} takes only $m+1$ distinct values the probabilities are heavily tied, so the threshold sweeps in whole tie blocks and coverage overshoots, in one cell from 0.593 to 0.873. Reported that way, every interval crosses zero.

Rescoring the stored records at exact coverage reverses that. Accepting exactly $k = \mathrm{round}(\text{coverage} \times n)$ items by descending predicted probability, with ties broken by a seeded uniform that never inspects the outcome and averaged over 100 draws, matches the baseline coverage in every cell to three decimals and gives:

\begin{table}[htbp]
\centering
\small
\begin{tabular}{lrrlr}
\toprule
cell & baseline coverage & FP reduction, exact $k$ & 95\% CI & as-run \\
\midrule
qwen7b-gsm8k & 0.593 & \textbf{$+1.4$ pp} & [$+0.3$, $+2.8$] & $+1.1$ \\
qwen7b-math500 & 0.110 & \textbf{$+15.5$ pp} & [$+2.7$, $+31.3$] & $+11.9$ \\
mistral7b-gsm8k & 0.704 & $+0.9$ pp & [$-2.0$, $+3.6$] & $+0.1$ \\
mistral7b-math500 & 0.333 & $+1.6$ pp & [$-5.1$, $+8.2$] & $+1.6$ \\
llama31-8b-gsm8k & 0.503 & \textbf{$+4.2$ pp} & [$+1.9$, $+7.0$] & $+0.7$ \\
llama31-8b-math500 & 0.152 & \textbf{$+14.0$ pp} & [$+1.9$, $+27.7$] & $+12.2$ \\
r1-distill-7b-gsm8k & 0.023 & $-6.4$ pp & [$-37.8$, $+25.7$] & $-6.4$ \\
r1-distill-7b-math500 & 0.015 & $+54.7$ pp & [$+0.0$, $+100.0$] & $+50.0$ \\
\bottomrule
\end{tabular}
\end{table}

H5 is supported on both registered anchor cells, at $+1.4$ and $+15.5$ points with intervals excluding zero, and the registration places the confirmatory H5 claim on the anchor. The same analysis clears the criterion on both Llama-3.1 cells, which is panel extension rather than confirmatory evidence, and neither Mistral cell reaches it, both being positive and not significant. The intervals are percentile bootstrap over 1000 problem-level resamples of the test split with the model held fixed, matching the as-run procedure.

Two rows must not be read as effects. The $+54.7$ point reduction on distilled MATH-500 rests on a baseline coverage of 0.015, roughly seven chains, with an interval whose lower bound is exactly zero, and the distilled GSM8K row at 2.3\% coverage is equally uninterpretable and happens to point the other way. Both are the degenerate-n pattern already reported for H4.

One row outside the registered panel belongs alongside these. On the exploratory cell of Section~\ref{sec:budget} the reduction is $+31.4$ points with an interval of [$+12.7$, $+52.4$]. It cannot support H5 and enters no tally. It sits on the cell with the cleanest reference chains in the study, at 3.2\% truncation against the registered cell's 36\%, which is a lead for anyone rerunning H5 rather than a result.

The tie rule was chosen after the as-run comparison was found to be mislabeled. Appendix~\ref{app:asrun} carries that comparison, the bootstrap and tie-break detail, and the headroom argument that bears on how a weak result here should be read.

\section{Protocol Dependencies}
\label{sec:dependencies}

\subsection{The Reference-Chain Token Budget}
\label{sec:budget}

The harness generates the reference chain at four times the continuation cap, so the registered 150-token continuation setting gives a 600-token reference budget. Whether that budget binds was measured rather than assumed, first on a 50-problem sample before the panel was rented and then on all 500 anchor MATH-500 problems afterwards.

Four reference-only scans over the same 500 problems, at budgets of 512, 600, 1280 and 4096, give truncation rates of 45.0\%, 36.0\%, 3.2\% and 0.4\%. The median chain length does not move across the first three, at 472, 470 and 471 tokens, and the ninetieth percentile converges by 1280, at 1019 against 1025 tokens. The budget therefore never touches the typical chain. It releases a censored tail, and the largest uncensored chain observed at 4096 is 2,390 tokens, so no budget tested here clears the anchor entirely.

\begin{figure}[htbp]
\centering
\includegraphics[width=\textwidth]{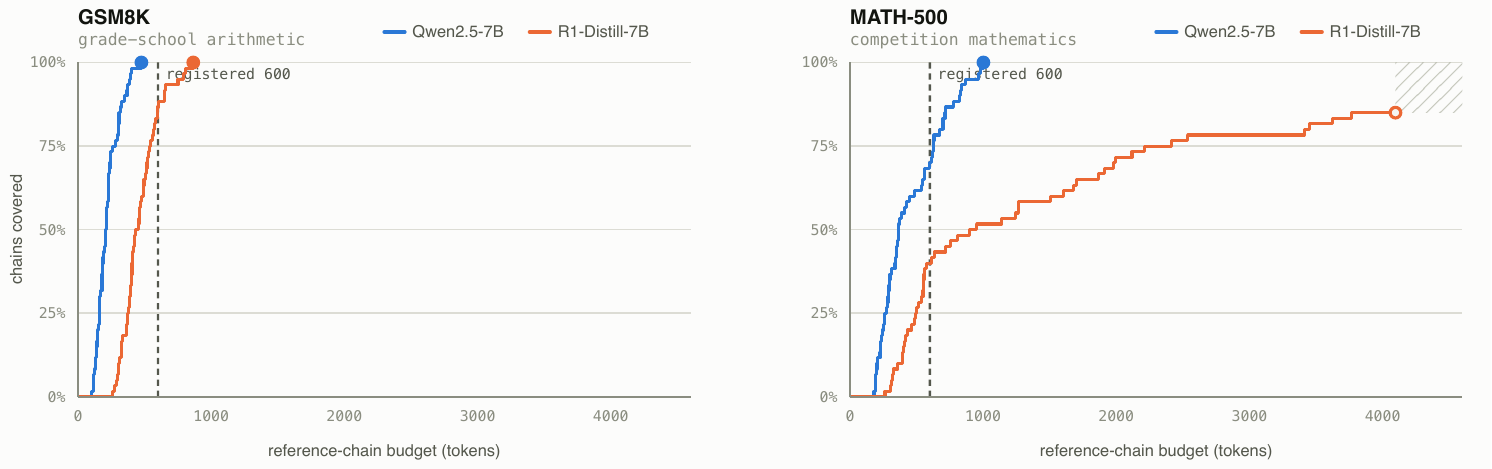}
\caption{Coverage curves: fraction of reference chains completing within a given token budget, by model and benchmark. The curves come from the 50-problem sample drawn for each model and benchmark before the panel ran, which puts the anchor's MATH-500 completion at 600 tokens at 70\%. The full 500-problem anchor scan gives 64.0\% at that budget, and its four truncation rates are reported in the text above.}
\label{fig:coverage}
\end{figure}

Where each curve crosses the dashed rule at 600 tokens is the fraction of chains that survive the protocol as registered: 100\% for the anchor on GSM8K, 86.7\% for the distilled model on GSM8K, 70\% for the anchor on MATH-500 and 40\% for the distilled model on MATH-500, from the 50-draw sample. The registered protocol is therefore measuring four materially different things across the four cells. Endpoint marks carry the censoring, filled where a curve reaches a real maximum and hollow where it stops at the 4096 probe ceiling. Chains recorded at exactly 4096 are right-censored rather than observed there, and are deliberately not drawn as observations, since doing so would carry the curve to 100\% and assert a measurement this study does not have.

The consequence for H2 is what the exploratory cell was built to test. The mechanism by which truncation could manufacture the observed positive $\rho$ is specific. A truncated chain has not converged, so its final entropy stays high and $C = H_0 - H_N$ is small, while truncation tracks chain length, which tracks difficulty, which tracks being wrong, so small C would pair with wrong answers. The anchor MATH-500 cell was therefore rerun with the reference budget raised from 600 to 1280 and every other registered parameter held fixed.

\begin{table}[htbp]
\centering
\small
\begin{tabular}{lrrrl}
\toprule
 & reference budget & measured truncation & magnitude $\rho$ & 95\% CI \\
\midrule
registered & 600 & 36.0\% & $+0.414$ & [$+0.342$, $+0.489$] \\
exploratory & 1280 & 3.2\% & $+0.497$ & [$+0.423$, $+0.568$] \\
\bottomrule
\end{tabular}
\end{table}

Rho moved away from zero rather than toward it, and the new interval's lower bound of $+0.423$ sits above the old point estimate. The two intervals overlap, so the increase itself should not be reported as significant, but the direction is unambiguous and the truncation explanation is not supported. The rest of the cell moved little, with accuracy 0.505 in both, monotone coverage 0.146 against 0.144, and the shape gap easing from $+27.49$ to $+25.38$ points. Two qualifications ride with the result. The two cells are independent generations at temperature 0.7 rather than the same chains measured twice, agreeing on only 47.6\% of predictions, so the direction of the change is interpretable and its magnitude is not, and the continuations were deliberately left at 150 tokens, since lengthening them would change what the entropy is computed over.

The original states a 512-token budget and reports no truncation rate. Reconstructed at that setting in this study's harness, the anchor censors 45.0\% of MATH-500 reference chains, so the published budget is tight enough for censoring on that scale to be possible and no rate is reported either way. Appendix~\ref{app:budget} carries that reconstruction, the distilled model's own rates, and a ceiling probe at 4096 tokens which leaves 22\% of its MATH-500 chains still truncated.

\subsection{Step Segmentation and the Step Cap}
\label{sec:segmentation}

The primary segmentation rule registered here is blank-line splitting capped at 8 units, applied uniformly to every model, and its behavior is not uniform across model families. Exclusions under the fewer-than-two-steps rule run from 0.0\% on both distilled cells and 0.2\% on Llama-3.1 GSM8K to 14.0\% on Mistral GSM8K, where 185 of 1319 problems yield fewer than two parseable steps. Two consequences follow, and both limit what cross-model comparisons can say. Monotonicity assessed over two or three points is a much weaker test than over five to nine, and short trajectories inflate coverage mechanically, which is part of why Mistral reads as 69.3\% monotone against the anchor's 56.3\%. Cross-model comparison of coverage is therefore not like-for-like, though the accuracy-gap comparison survives.

Zhao reports zero excluded steps across 1,474 nominal steps at $n=300$ under his marker-regex rule, so the exclusions here are a property of this study's segmentation choice rather than of the published method. A traced run of 100 Mistral GSM8K problems identifies the mechanism, and it is less flattering than model-dependence. Of the 14 chains excluded in that sample, 13 have fewer than two blank-line units, and all 14 reach two or more units under the pre-specified newline rule. The system prompt instructs the model to put each reasoning step on its own line, while the segmenter requires a blank line between steps, so a model that follows the instruction literally produces text the segmenter reads as a single unit. Qwen inserts blank lines and loses 0.2\%; Mistral does not and loses 14.0\%. The mismatch under-segments everywhere rather than only where it excludes, at a median unit count under the registered rule against the newline rule of 2 against 8 for Mistral and 7 against 25 for the anchor, so the trajectories the panel is built on are coarser than the reasoning they measure. This is a defect in the protocol as registered here. What the rule does establish is that a reasonable step-segmentation choice produces exclusion rates from 0.0\% to 14.0\% across four models, which suggests the protocol's step-boundary requirement is more model-dependent than a single-family evaluation would reveal. The sensitivity analysis that would settle it was pre-specified and has not been run.

The cap of 8 units has no counterpart in the original, whose chains average 4.9 steps. It is a cost-control choice made here, and it binds hard on long chains. The reference-only scans record raw unit counts with a median of 7 to 8 and a maximum of 55, and 46.4\% to 55.6\% of anchor MATH-500 chains carry at least 8 units depending on the budget. For those chains the last trajectory point merges an unbounded tail of reasoning into a single prefix, which makes the final entropy not comparable with the final entropy of a chain that fits. Reporting the fraction of chains that reach the cap is therefore necessary, and is done here.

A second consequence of the same interaction affects what the final entropy means. When the last prefix of a chain returns nothing, the final trajectory point is the prefix before it, so the final entropy is measured over the chain minus the unit that states its answer. In the traced sample that is the normal case for a chain that finished, at 65 of 67 complete chains against none of the 33 truncated ones. It is therefore not the same quantity across the two groups, and because coherence is the first-step entropy minus the final one, the magnitude measure inherits the asymmetry. Whether it moves the result of Section~\ref{sec:magnitude} is untested, and it is not the kind of asymmetry a rank correlation ignores, since the omitted unit is the one at which entropy would be lowest and it is omitted from one group and not the other. Settling it means recomputing the trajectory statistics from retained continuation text, which the panel did not keep. Appendix G records a related artifact in the trajectory-length counts.

\section{Secondary Analyses}
\label{sec:secondary}

\subsection{Scalar Baselines and Confounder Controls}
\label{sec:scalar}

The analyses in this section are exploratory and outside the registration. They reproduce two checks the original performs on its own result, and both bear on how the study's confirmatory findings should be read.

The first is the original's confounder control. Residualizing both the monotonicity flag and correctness on trajectory length and correlating the residuals gives a partial correlation that is positive and significant in six of the eight cells, running from $+0.110$ to $+0.272$, against the $+0.179$ the original reports on its pilot. The two exceptions are mistral7b-math500 at $+0.008$ and r1-distill-7b-math500 at $+0.028$, neither near significance. The shape effect therefore survives adjustment for chain length in the cells where it is present at all, which is the same conclusion the original reaches. Trajectory length is an imperfect stand-in for the chain-length variable the original uses, since it counts prefixes that yielded an extractable answer rather than segmentation units, so the adjustment is approximate.

The second is the comparison against cheap scalar baselines, and it does not reproduce. Ranking chains by the negated final-step entropy discriminates correctness better than the binary monotonicity flag in all eight cells by area under the ROC curve. By area under the risk-coverage curve, the statistic the original reports, it does so in seven cells when the area is taken over the full curve and in six when it is truncated at the original's 73.7\% coverage, and the original does not state which range it uses. The exception under both is qwen7b-gsm8k, which is the cell configured closest to the original's, and llama31-8b-gsm8k joins it under truncation. Two qualifications belong with that before it is read as a refutation. The ROC area is not the original's statistic, since its table reports selective accuracy at a fixed 73.7\% coverage beside a risk-coverage area, and a binary flag is at a structural disadvantage on any ranking measure because it takes two values and cannot order chains within either group. The risk-coverage recomputation answers the first. The comparison that addresses the second is against the violation count, which is also a trajectory-shape measure but takes many values. It beats the binary flag in all eight cells by ROC area, as expected, and is itself beaten by final-step entropy in seven of eight by ROC area and in six by risk-coverage area under either range. Appendix~\ref{app:scalar} carries the ranking tables and a logistic regression that agrees on which term carries the weight.

This is the same phenomenon as the H2 failure rather than a separate one. Coherence is $C = H_0 - H_N$, so a final entropy that predicts correctness strongly makes C predictive too, and the cells where coherence discriminates best are exactly the cells where H2 fails hardest. What the two results say together is that in these runs the informative quantity is where the trajectory ends, not whether it descended monotonically to get there.

Both carry a dependency the study can size but not remove. The correctness label is the majority answer of the last prefix's five samples, and the final entropy is the entropy of that same five, so a low final entropy is in part a statement that the scored answer had consensus behind it. The traced run of Appendix~\ref{app:traced} retains the reference chain, which allows the label to be recomputed from the chain's own answer instead. On its anchor MATH-500 arm the two labels agree on 85\% of items, and decoupling lowers the final-entropy discrimination from 0.772 to 0.698 while leaving the monotonicity and violation-count rows unmoved at 0.541 and 0.660. The same recomputation moves the coherence association from $+0.350$ to $+0.315$ on that arm, which is why Section~\ref{sec:magnitude}'s failures are not attributed to this dependency.

The third check reproduces the original's other magnitude analysis, and it is the one that survives. Zhao reports that among non-monotone chains the size of the largest upward entropy step is unassociated with correctness, at $\rho$ $-0.017$ with p=0.88, which he offers as an extension of the shape-over-magnitude dissociation to the multi-violation setting. Recomputing that statistic on each cell's stored trajectories:

\begin{table}[htbp]
\centering
\small
\begin{tabular}{lrrr}
\toprule
cell & non-monotone $n$ & $\rho$(max positive $\Delta H$, correctness) & $p$ \\
\midrule
qwen7b-gsm8k & 566 & $-0.044$ & 0.292 \\
qwen7b-math500 & 426 & $-0.105$ & 0.030 \\
mistral7b-gsm8k & 348 & $+0.028$ & 0.598 \\
mistral7b-math500 & 319 & $-0.028$ & 0.615 \\
llama31-8b-gsm8k & 644 & $-0.073$ & 0.063 \\
llama31-8b-math500 & 415 & $-0.025$ & 0.609 \\
r1-distill-7b-gsm8k & 1302 & $-0.053$ & 0.055 \\
r1-distill-7b-math500 & 497 & $-0.144$ & 0.001 \\
\bottomrule
\end{tabular}
\end{table}

Violation magnitude is null or negligible in every cell, crossing p=0.05 in two of eight at $\rho$ $-0.105$ and $-0.144$, while the violation count over the same records is significant in all eight at $-0.108$ to $-0.338$. The original's dissociation therefore replicates in its violation form across four models and both benchmarks, on a panel more than twenty times the size of the one it was reported on. How many disruptions a trajectory contains predicts correctness; how large they are adds little beyond it. The result is exploratory, since the registration names no hypothesis about violation magnitude, and it is the reason this paper does not report the dissociation as failing wholesale.

\section{Limitations}
\label{sec:limitations}

The reference-chain temperature is the limitation most likely to matter, and it is now partly measured. The original generates its chain at temperature 0.1 and this study generates it at 0.7. The concern was that a hotter chain is longer, less likely to converge, and more likely to be truncated, and that all three would inflate the final entropy on exactly the problems the model gets wrong, which would manufacture the observed association between coherence and correctness.

Two of those three consequences were tested with a pair of reference-only scans at a 512-token budget differing only in temperature, and they are absent. Truncation is 45.0\% at temperature 0.7 and 44.8\% at 0.1, median chain length 471 against 476 tokens, and the share of chains reaching the segmentation cap 40.6\% against 42.2\%. That the temperature was genuinely applied is not in doubt. Among the 246 problems where both arms produced a complete chain only 2.8\% have identical token counts, and 195 of roughly 225 truncated problems are truncated in both arms, so which problems overrun the budget appears to be a property of the problem rather than of the sampling temperature. What remains untested is convergence, which is a claim about where a chain ends rather than how long it runs. The coherence failures should therefore be read as holding under this study's sampling regime, with the length and truncation routes excluded and the semantic route open. Running the anchor cell at temperature 0.1 would close it and remains the recommended next measurement.

The segmentation rule and the step cap are likewise this study's choices, and Section~\ref{sec:segmentation} shows they produce exclusion rates and merged tails that the original does not report and would not necessarily incur. The pre-specified segmentation sweep would bound this and has not been run.

A dependency in the outcome definition sits upstream of every magnitude result and is only partly measured. Correctness is the majority of the last prefix's five sampled answers, and the final entropy is computed from those same five, so the two share a draw, and the reference chain's own answer is never scored. Section~\ref{sec:scalar} recomputes the label from the chain on a traced subsample and finds the coherence association essentially unmoved and the final-entropy discrimination materially lower. That bounds the problem on 100 problems of one cell, in one generation, and does not settle it at panel scale, since a rescoring against a chain-derived label is possible only where the chain was retained. Until then the magnitude results should be read as holding under this study's outcome definition, with the direction of the dependency measured and its size at scale unknown.

Single-seed operation applies throughout. The registration pre-specifies a three-seed check as secondary and it was not run, so run-to-run variation is largely unmeasured, though the original's own three-seed sweep on GSM8K spans $+5.2$ to $+13.8$ points and suggests it is not small. One partial measurement exists. The two anchor MATH-500 cells agree per problem on 47.6\% of predictions, 76.1\% of correctness, 83.3\% of monotonicity flags and 3.0\% of trajectories. They differ in reference budget as well as in sampling, so this bounds agreement between two generations rather than isolating seed variation. Across these two temperature-0.7 generations only 3.0\% of trajectories match exactly, which shows low trajectory-level agreement even between two nearby configurations.

The false-positive-reduction result in Section 4.5 rests on a rescoring at exact coverage rather than on the as-run analysis, whose thresholding overshot its target. The tie rule is outcome-independent and averaged over draws, but it is a choice made after seeing that the as-run comparison was mislabeled, and two of its eight cells rest on degenerate counts. H4's estimand is undefined in practice on the model it was designed for, so the study reports a limit of the method and not a measurement of the effect, and on one model rather than on the class it was drawn from. Coverage of models is four, all in the 7-to-8-billion-parameter range, and coverage of tasks is two mathematics benchmarks with extractable numeric or boxed answers, so nothing here speaks to open-domain reasoning, where answer canonicalization is the harder problem.

Finally, a signal of this kind invites the response of training against it. A model tuned to produce smoothly decreasing answer entropy would satisfy the monotonicity criterion without becoming more reliable, so claims from this line of work should stay scoped to diagnosis and should not be offered as certification.

\section{Conclusions}
\label{sec:conclusions}

The entropy-trajectory shape signal replicates cleanly on the three standard instruct models, across both complete test sets. Over the full four-model panel the gap interval excludes zero in seven of eight cells, though that count should not be read as validation on the fourth model, where monotone coverage collapses to about one chain in a hundred and the eighth cell has no power. The graded violation count predicts correctness in all eight. On the anchor model the effect sizes are close to the original's own full-test-set figures, and the agreement holds across a different segmentation rule, a different reference temperature and a different decoding stack, which makes it a statement about robustness rather than about pipeline agreement.

The shape-over-magnitude dissociation divides in two, and the halves behave differently. Its violation half replicates. Among non-monotone chains the size of the largest entropy rise is null or small in all eight cells, two of them nominally significant at $\rho$ $-0.105$ and $-0.144$ and both far below the violation count, which is significant in all eight. That reproduces the original's own multi-violation result on a panel more than twenty times larger. Its scalar-coherence half is setting-dependent. Trajectory magnitude meets the registered null criterion in two of the six standard-model cells that are its registered scope, and the two distilled cells also exceed the threshold. One of the two cells that meet it is the single cell matching the conditions under which the original measured coherence at all, and there the null reproduces at $\rho$ $-0.018$ against its $-0.059$. Every cell that misses the criterion is a condition for which the original publishes no coherence figure, so those results extend the map rather than contradict a reported point.

Two candidate explanations for those results find no support and two remain open. Raising the reference-chain token budget, which cut measured truncation from 36.0\% to 3.2\%, moved the correlation further from zero rather than toward it, so this study's own reference-chain truncation is not supported as the explanation. Reference-chain temperature was tested on the routes by which it could act. Sampling at the original's temperature rather than this study's changes every individual chain and leaves the distribution of lengths and truncation rates unmoved, with the same problems truncating in both cases. What remains open is whether equal-length chains converge differently, which no scan can reach, and the endpoint asymmetry of Section~\ref{sec:segmentation}. In the traced Qwen MATH-500 subsample, recomputing the outcome label without its shared sample leaves the coherence association substantial, so same-sample coupling does not fully explain that cell's result, and its contribution across the panel is unknown.

One exploratory result bears on how the two confirmatory ones fit together. By ROC area, ranking chains by the final-step entropy alone discriminates correctness better than the binary monotonicity flag in all eight cells and better than the graded violation count in seven. Under the risk-coverage area the original reports, the comparison with the binary flag holds in six or seven cells, depending on an integration range the original does not specify. That is not a second finding so much as the first one seen from another angle, since coherence is the first-step entropy minus the final one, and it suggests that in these runs the informative quantity is where a trajectory ends rather than whether it descended monotonically to get there. How broadly it holds depends on the statistic, and a binary flag is disadvantaged on any ranking measure, so it is offered as a direction to test rather than as a settled ordering.

Applied to a reasoning-distilled model the protocol yields a predictor that fires on about one chain in a hundred, trajectories that run against the step cap, and an accuracy figure that reflects the token budget rather than the model. The registered contrast for that model is therefore reported as undefined in practice. This appears to be the more general result of the study. The method's applicability is bounded by properties of the chains a model emits, and those properties varied enough across the four models tested here that a protocol validated on one should not be assumed to transfer.

Three recommendations follow for anyone building on this line of work. Reporting the reference-chain truncation rate alongside any trajectory result is the foremost one, since a third of hard-problem chains were censored at a budget close to the published setting and nothing in a trajectory statistic reveals it. Reporting the fraction of chains that reach the step cap is recommended for the same reason. And a sweep over the number of samples per step would be a more informative robustness check than a sweep over the monotonicity tolerance, whose insensitivity at $m=5$ is close to arithmetically determined.

\section*{Reproducibility and Data Availability}

The harness, configurations, registered plan and the panel analysis scripts are public at \url{https://github.com/ai4altruism/entropy-drift-pilot}. The code for the additional analyses in Section~\ref{sec:scalar} and in Appendices~\ref{app:earlyexit}, \ref{app:scalar} and \ref{app:tolerance}, and for the multiplicity correction in Appendix~\ref{app:holm} is held in the author's private working repository and will be added to the public one on release; it imports its scoring functions from the public harness rather than reimplementing them, so the definitions applied there are the registered ones. The preregistration is frozen at commit \texttt{a2edb8e9}, SHA-256 \texttt{036288be...873cd}, reachable as tag \texttt{prereg-v1.0}, and registered at OSF as \texttt{osf.io/8w2q3}, \texttt{doi:10.17605/OSF.IO/8W2Q3}. The registration is public, and its embargo was ended early on 2026-09-16. The eight confirmatory cells ran at harness commit \texttt{ab5c6968} and the exploratory cell at \texttt{1eb6b562}. Every cell writes a provenance manifest containing the harness commit, the resolved model commit SHA, the GPU, the CUDA and torch versions and the package set. Record files, manifests and per-cell summaries for all nine cells, together with the three reference-only scans, are retained in four verified copies.

Models and datasets are artifacts rather than papers, and are identified by Hugging Face repository and by the revision pinned at run time, as recorded in Section~\ref{sec:panel} and in the run manifests. \texttt{Qwen/Qwen2.5-7B-Instruct} at \texttt{a09a3545} is described by \citet{qwen2024}, \texttt{meta-llama/Llama-3.1-8B-Instruct} at \texttt{0e9e39f2} by \citet{grattafiori2024llama3}, and \texttt{deepseek-ai/DeepSeek-R1-Distill-Qwen-7B} at \texttt{916b56a4} by \citet{deepseek2025r1}, which introduces the distilled models alongside R1 itself. \texttt{mistralai/Mistral-7B-Instruct-v0.3} at \texttt{c170c708} is the exception. \citet{jiang2023mistral} describes the base model and the v0.1 instruct release rather than these weights, so the model card is the primary record for what was run. \texttt{openai/gsm8k} is the dataset of \citet{cobbe2021gsm8k}. \texttt{HuggingFaceH4/MATH-500} states on its own card that it is the 500-problem subset of the MATH benchmark of \citet{hendrycks2021math} that OpenAI drew for \citet{lightman2023verify} and distributes in the \texttt{openai/prm800k} repository; that split, and not the full MATH test set, is what MATH-500 means throughout this paper.

\bibliography{refs}

\appendix

\section{The As-Run False-Positive Comparison}
\label{app:asrun}

Section~\ref{sec:triage} reports H5 at the exact coverage the registration specifies. This is the comparison the as-run code produced instead, at each cell's own overshot coverage, retained so the repair is auditable against what it replaced.

\begin{table}[htbp]
\centering
\small
\resizebox{\textwidth}{!}{%
\begin{tabular}{lrlll}
\toprule
cell & FP reduction & 95\% CI & baseline $\rightarrow$ filter coverage & baseline $\rightarrow$ filter sel. accuracy \\
\midrule
qwen7b-gsm8k & $+1.1$ pp & [$-1.2$, $+3.6$] & 0.593 $\rightarrow$ 0.873 & 0.876 $\rightarrow$ 0.887 \\
qwen7b-math500 & $+11.9$ pp & [$-4.7$, $+30.9$] & 0.110 $\rightarrow$ 0.325 & 0.727 $\rightarrow$ 0.846 \\
mistral7b-gsm8k & $+0.1$ pp & [$-3.2$, $+3.4$] & 0.704 $\rightarrow$ 0.740 & 0.533 $\rightarrow$ 0.534 \\
mistral7b-math500 & $+1.6$ pp & [$-5.8$, $+8.6$] & 0.333 $\rightarrow$ 0.333 & 0.115 $\rightarrow$ 0.131 \\
llama31-8b-gsm8k & $+0.7$ pp & [$-2.5$, $+4.3$] & 0.503 $\rightarrow$ 0.715 & 0.906 $\rightarrow$ 0.912 \\
llama31-8b-math500 & $+12.2$ pp & [$-4.3$, $+28.3$] & 0.152 $\rightarrow$ 0.228 & 0.500 $\rightarrow$ 0.622 \\
r1-distill-7b-gsm8k & $-6.4$ pp & [$-35.2$, $+26.2$] & 0.023 $\rightarrow$ 0.025 & 0.833 $\rightarrow$ 0.769 \\
r1-distill-7b-math500 & $+50.0$ pp & [$+0.0$, $+100.0$] & 0.015 $\rightarrow$ 0.030 & 0.333 $\rightarrow$ 0.833 \\
\bottomrule
\end{tabular}}
\end{table}

Every interval crosses zero. The coverage columns show why this is the wrong comparison. The filter accepts well past the baseline's coverage in five cells. The selective-accuracy columns show it is nonetheless more accurate on what it accepts in seven of the eight, which is the dominance property Section~\ref{sec:triage} describes.

The intervals incorporate the tie-breaking variation as well as the resampling, since each bootstrap replicate re-draws its own tie-breaks, though it averages only five of them against the hundred behind each point estimate, so a small part of the interval width is residual tie-break noise rather than sampling uncertainty. The two distilled cells remain uninterpretable, and the MATH-500 one does not meet the criterion in any case, since an interval whose lower bound is exactly zero does not exclude it.

The direction of the exact-coverage correction was predictable. The overshoot admits more items, which normally costs selective accuracy, so it worked against the learned filter. What was not predictable was the size. On the two cells where the filter now clears the criterion most comfortably the point estimate barely moved, while on llama31-8b-gsm8k it moved from $+0.7$ to $+4.2$ points and from an interval crossing zero to one excluding it. A registered hypothesis was reported as a null for want of a tie rule.

The as-run comparison is retained because it says something the exact-coverage test does not. At its own overshot coverage the filter accepts more items and is more accurate on what it accepts in seven of the eight cells, the exception being distilled GSM8K. That is dominance on both axes, a stronger property than beating a baseline at one operating point, and it is visible in the full coverage curve rather than at a single k.

One further consideration bears on how a weak result should be read. The design assumed a monotone-but-wrong rate near the original's 31.2\%, and on the anchor's GSM8K cell the observed rate is 11.2\%, so a filter has little room to improve on the bare flag there. The rate is dataset-dependent rather than uniformly low, at 26.0\% on the anchor's MATH-500 cell and 50.6\% on Llama-3.1's, so the ceiling explanation covers some cells and not others. Where headroom exists the point estimates are larger, at $+11.9$ and $+12.2$ points, and their intervals are correspondingly wide.

\section{Deviations and Incidents}
\label{app:deviations}

Four departures are recorded, all logged contemporaneously rather than reconstructed.

\textbf{The registered multiplicity correction.} Holm-Bonferroni was registered and not implemented in the as-run code. Section~\ref{sec:prereg} applies it to the family the frozen registration specifies. All four $p$-value-bearing tests remain significant and H5 meets its registered interval criterion on both anchor cells, so no registered verdict changes. The sixteen-test panel-wide correction reported in an earlier revision of this paper is retained there only as an exploratory robustness check.

\textbf{A duplicated record on crash-resume.} The \texttt{mistral7b-math500} cell was killed at index 11 with no traceback and, on resume, reprocessed that index because the reader did not see the record already on disk, finishing with 501 lines for 500 problems. The two copies are not identical, differing in coherence, violation count and predicted answer, so \texttt{seed: 0} does not make vLLM bit-reproducible and a duplicate is two genuine samples entering every statistic. The handling was fixed before its effect was measured: keep the first occurrence, a rule that never inspects values, so the recovery could not become a choice between two results. The effect was negligible, moving the gap from $+7.52$ to $+7.50$ points. The raw file is preserved. All other cells hold exactly distinct indices.

\textbf{A record that could not be regenerated.} Index 430 of \texttt{llama31-8b-gsm8k} was destroyed by a mid-write kill. The regeneration attempt was refused by the harness's resume check, which compares a SHA-256 of the serialized config: the repository had since gained an optional \texttt{reference\_max\_tokens} field, so the hash differed although no registered parameter had moved. The refusal is correct behavior for a check that cannot distinguish a cosmetic schema addition from a substantive change. That cell is reported at $n=1318$, and one record in 1318 cannot move its statistics materially. The general lesson is that updating code mid-study can lock a run out of resuming.

\textbf{One panel interruption.} The same kill stopped the driver after three of eight cells, because it used \texttt{set -e} and a wrapper printed a completion message regardless of exit status. No records were lost, and \texttt{--resume} continued the affected cell. The driver now isolates failures per cell and exits non-zero.

\section{The Registered Multiplicity Correction}
\label{app:holm}

\begin{table}[htbp]
\centering
\small
\begin{tabular}{p{0.05\textwidth}p{0.18\textwidth}p{0.24\textwidth}p{0.24\textwidth}p{0.14\textwidth}}
\toprule
 & registered scope & estimand & success rule & multiplicity family \\
\midrule
H1 & anchor, both datasets & accuracy gap, Fisher one-sided, odds ratio & gap CI excludes 0 positively on both datasets & primary, Holm \\
H2 & anchor, extended to the other standard models as secondary & Spearman $\rho(C, \text{correct})$ & absolute $\rho$ below 0.10 and substantially below the shape effect & not in the null-testing family \\
H3 & as H2 & $\rho(\text{violations}, \text{correct})$, bucket accuracies & $\rho$ below 0, CI excludes 0 & primary, Holm \\
H4 & distilled model & gap and its CI, and the difference from the mean gap of the three standard models & estimation, no criterion, two-sided by design & excluded from the family \\
H5 & learned filter on a 60/40 split & coverage-matched FP-rate difference, plus the coverage curve & reduction CI excludes 0 favorably, on the test split & primary, Holm \\
\bottomrule
\end{tabular}
\end{table}

Holm-Bonferroni was registered and was not implemented in the as-run analysis code, and is applied here over the family the registration names: H1 on both anchor datasets, H3, and H5. The registration does not say whether H3 counts as one test or as one per dataset, and it is counted here as one per dataset, which is the conservative reading because it raises $m$ and tightens every threshold. That gives a family of five, of which four carry $p$-values:

\begin{table}[htbp]
\centering
\small
\resizebox{\textwidth}{!}{%
\begin{tabular}{llllll}
\toprule
test & raw $p$ & Holm rank & threshold at that rank & Holm-adjusted $p$ & verdict \\
\midrule
H3, anchor MATH-500 & 8.2e-15 & 1 of 4 & 0.0125 & 3.3e-14 & reject \\
H3, anchor GSM8K & 8.4e-10 & 2 of 4 & 0.0167 & 2.5e-09 & reject \\
H1, anchor GSM8K & 1.6e-06 & 3 of 4 & 0.0250 & 3.2e-06 & reject \\
H1, anchor MATH-500 & 9.2e-06 & 4 of 4 & 0.0500 & 9.2e-06 & reject \\
H5, anchor both datasets & n/a & evaluated separately & n/a & n/a & interval criterion met \\
\bottomrule
\end{tabular}}
\end{table}

All four comfortably clear their thresholds, the closest being H1 on anchor MATH-500 at roughly three orders of magnitude. H5 is not ranked among them, because Holm is a procedure on ordered $p$-values and H5's registered criterion is an interval; it is evaluated separately against that criterion, which it meets on both anchor cells. Correcting conservatively over five family members instead, on the reading that H5 occupies a slot, moves every threshold down one rank and changes no verdict, so the bookkeeping choice is immaterial to every conclusion here. No registered verdict changes under the correction, which is why the omission is reported as a deviation rather than as a defect.

An earlier revision of this paper applied Holm to a sixteen-test panel-wide family of H1 and H3 in every cell. That family is not the registered one, it silently promoted secondary replication cells to primary status, and it omitted H5, which the registration includes. The panel-wide correction is retained as an exploratory robustness check, where it rejects fifteen of sixteen and retains only H1 on the distilled MATH-500 cell at $p=0.551$, the cell Section~\ref{sec:shape} already reports as powerless.

\section{Early Exit}
\label{app:earlyexit}

\label{sec:earlyexit}

The original reports that monotonicity computed over only the first two entropy transitions recovers 76\% of the full accuracy gap at 60\% of the trajectory cost, which would make an early-exit deployment attractive. Recomputing the same statistic on each registered cell, truncating each trajectory to its first $k+1$ points, does not reproduce that economy anywhere.

\begin{table}[htbp]
\centering
\small
\begin{tabular}{lrrrrr}
\toprule
cell & $k=1$ & $k=2$ & $k=3$ & full & $k=2$ as \% of full \\
\midrule
qwen7b-gsm8k & $+2.40$ & $+4.85$ & $+7.20$ & $+9.62$ & 50\% \\
qwen7b-math500 & $+1.65$ & $+5.96$ & $+14.24$ & $+27.49$ & 22\% \\
mistral7b-gsm8k & $+10.42$ & $+17.55$ & $+20.95$ & $+20.74$ & 85\% \\
mistral7b-math500 & $+0.10$ & $+3.89$ & $+2.59$ & $+7.50$ & 52\% \\
llama31-8b-gsm8k & $+7.33$ & $+12.24$ & $+17.23$ & $+21.86$ & 56\% \\
llama31-8b-math500 & $+2.72$ & $+9.76$ & $+15.48$ & $+21.88$ & 45\% \\
r1-distill-7b-gsm8k & $+3.37$ & $-1.14$ & $+0.42$ & $+48.94$ & $-2$\% \\
r1-distill-7b-math500 & $+3.73$ & $+4.68$ & $+0.47$ & $+9.99$ & 47\% \\
\bottomrule
\end{tabular}
\end{table}

The recovered fraction runs from $-2$\% to 85\% and reaches 76\% in no cell. On the anchor it is 50\% on GSM8K and 22\% on MATH-500. The ordering is informative. The fraction is highest where trajectories are shortest, at 85\% on the cell with a median of 3 points, where truncating at $k=2$ often removes nothing, and lowest on the long-chain cells, which is what one would expect if the gap accumulates across the chain rather than sitting in its opening. On the distilled GSM8K cell the early-exit gap is negative at $k=2$ while the full gap is $+48.94$ points, so on that cell the signal lives in the tail and truncation destroys it rather than cheapening it.

The comparison is not at matched coverage, and the direction of the error is known. Truncating a trajectory removes chances to violate, so the flag fires more often on a prefix, and coverage rises from 0.563 to 0.656 on the anchor's GSM8K cell and from 0.146 to 0.385 on its MATH-500 cell. Part of every reduction above is therefore a move to a different operating point rather than lost signal. The original's prefix table has the same structure, with coverage of 82.4\% at $k=2$ against 73.7\% at full trajectory, and its subset-shift control does not address coverage. The reading that fits both is that early exit trades a smaller gap for higher coverage, and that neither paper's recovered-fraction number summarizes that trade.

\section{The Reference-Chain Budget in Detail}
\label{app:budget}

Zhao specifies a 512-token budget, slightly tighter than the 600 used here, and reports no truncation rate. Reconstructed at that setting in this study's harness, the anchor censors 45.0\% of MATH-500 reference chains. That figure describes this reconstruction and not the original's own execution, which cannot be observed without its outputs or its code, neither of which is available; what it establishes is that the published budget is tight enough for censoring on that scale to be possible, and that no rate is reported either way. That is a gap in the original's reporting rather than in its specification, and it is larger at the original's own budget than at this study's. It is a gap worth closing, because three of this study's cells would otherwise be uninterpretable. On a 50-problem sample taken before the panel was run, the distilled model truncated at 18\% on GSM8K and 58\% on MATH-500 at a 600-token budget, which is why its accuracy figures measure the instrument. Those two rates carry the sampling caveat of Section~\ref{sec:budget}, since the same 50-draw method understated the anchor's MATH-500 rate as 30\% where the full scan gives 36.0\%.

The same sample also probed the ceiling, and that probe is what distinguishes the distilled model from a model that merely needs a larger budget. Raising the reference budget to 4096 and re-running the same 50 MATH-500 problems leaves 22\% of its chains still truncated, with the ninetieth percentile onward sitting on the cap, so the location of its tail is unknown at every budget tested here. Its lengths are bimodal in that sample: 22 of 50 chains finish under 750 tokens while 11 run past 4000, which is a model that either solves quickly or enters an extended deliberation no budget tested terminates. The anchor at the same ceiling has a longest chain of 1004 tokens in 50 draws, and on all 500 problems a residual of 0.4\%. The distinction matters for how Section~\ref{sec:distilled} should be read. The protocol's assumption that a reasoning chain fits a bounded token budget is one this model violates structurally rather than marginally, and that is a stronger and more general statement than its 600-token rate alone supports. Both distilled figures rest on 50 draws, and the anchor's own 50-draw maximum understated its full-scan tail by more than half, so the direction is well supported and the magnitudes are not.

\section{Scalar Baselines}
\label{app:scalar}

This appendix carries the ranking comparisons Section~\ref{sec:scalar} summarizes, first by ROC area and then by risk-coverage area, and the logistic regression that agrees with both.

\subsection{ROC Area}

\begin{table}[H]
\noindent Area under the ROC curve, higher being better:\par\medskip
\centering
\small
\begin{tabular}{lrrrr}
\toprule
cell & monotonicity flag & violation count & final-step entropy & coherence $C$ \\
\midrule
qwen7b-gsm8k & 0.591 & 0.620 & 0.606 & 0.486 \\
qwen7b-math500 & 0.569 & 0.690 & 0.771 & 0.738 \\
mistral7b-gsm8k & 0.589 & 0.596 & 0.612 & 0.517 \\
mistral7b-math500 & 0.601 & 0.641 & 0.701 & 0.668 \\
llama31-8b-gsm8k & 0.667 & 0.700 & 0.728 & 0.589 \\
llama31-8b-math500 & 0.568 & 0.675 & 0.774 & 0.730 \\
r1-distill-7b-gsm8k & 0.514 & 0.562 & 0.812 & 0.779 \\
r1-distill-7b-math500 & 0.502 & 0.599 & 0.786 & 0.758 \\
\bottomrule
\end{tabular}
\end{table}

The ROC area is not the original's statistic, since the original's table reports selective accuracy at a fixed 73.7\% coverage beside a risk-coverage area. A binary flag is also at a structural disadvantage on any ranking measure, because it takes two values and cannot order chains within either group. The comparison that addresses the second objection is against the violation count, which is also a trajectory-shape measure but takes many values. It beats the binary flag in all eight cells, as expected, and is itself beaten by final-step entropy in seven of eight. The single exception is the anchor model on GSM8K, at 0.620 against 0.606, which is the cell whose configuration is closest to the original's.

\subsection{Risk-Coverage Area}

Recomputing the comparison under the risk-coverage statistic the original reports answers the first objection. AURC is the area under the risk-coverage curve, lower being better. The original does not state the range its area is taken over, and the clues in its own table point in different directions. It reports the same AURC of 0.311 for the binary flag and the violation count. That is what an area stopping at the flag's 73.7\% coverage would give, since both signals rank the same 221 monotone chains first. A full-curve area would generally separate them, because the violation count also orders the 79 non-monotone chains the flag leaves tied. Its oracle AURC of 0.068 is hard to reconcile with an area stopping at 73.7\%, on which a perfect ranking of its 300 problems at 63.0\% accuracy would score about 0.01. Rather than reconstruct an undocumented implementation, the area is computed both ways here. Over the full curve it is the mean selective risk at every coverage level from $1/n$ to 1, and truncated it is the same mean over coverage levels up to 0.737 only. Ties are broken by a seeded uniform averaged over 50 draws, the same draws for both ranges, so that the heavily tied binary flag is not decided by sort order.

\begin{table}[H]
\noindent Over the full curve:\par\medskip
\centering
\small
\resizebox{\textwidth}{!}{%
\begin{tabular}{lrrrrr}
\toprule
cell & full-coverage accuracy & monotonicity flag & violation count & final-step entropy & coherence $C$ \\
\midrule
qwen7b-gsm8k & 0.846 & 0.123 & \textbf{0.119} & 0.127 & 0.161 \\
qwen7b-math500 & 0.505 & 0.417 & 0.358 & \textbf{0.330} & 0.342 \\
mistral7b-gsm8k & 0.466 & 0.483 & 0.481 & \textbf{0.474} & 0.530 \\
mistral7b-math500 & 0.085 & 0.885 & 0.883 & \textbf{0.875} & 0.880 \\
llama31-8b-gsm8k & 0.793 & 0.131 & 0.123 & \textbf{0.126} & 0.185 \\
llama31-8b-math500 & 0.309 & 0.626 & 0.582 & \textbf{0.523} & 0.553 \\
r1-distill-7b-gsm8k & 0.340 & 0.631 & 0.594 & \textbf{0.436} & 0.468 \\
r1-distill-7b-math500 & 0.234 & 0.765 & 0.715 & \textbf{0.576} & 0.580 \\
\bottomrule
\end{tabular}}
\end{table}

AURC is bounded by a cell's base error rate, so these compare within a row and not across rows. Over the full curve the ordering survives the change of statistic. The violation count beats the binary flag in all eight cells and the final-step entropy beats it in seven, the exception being qwen7b-gsm8k, which is again the cell configured closest to the original's.

\begin{table}[H]
\noindent Truncated at 73.7\% coverage:\par\medskip
\centering
\small
\begin{tabular}{lrrrr}
\toprule
cell & monotonicity flag & violation count & final-step entropy & coherence $C$ \\
\midrule
qwen7b-gsm8k & \textbf{0.114} & 0.115 & 0.127 & 0.163 \\
qwen7b-math500 & 0.392 & 0.322 & \textbf{0.296} & 0.309 \\
mistral7b-gsm8k & 0.473 & 0.473 & \textbf{0.467} & 0.532 \\
mistral7b-math500 & 0.876 & 0.874 & \textbf{0.865} & 0.871 \\
llama31-8b-gsm8k & 0.110 & \textbf{0.105} & 0.119 & 0.186 \\
llama31-8b-math500 & 0.605 & 0.553 & \textbf{0.477} & 0.516 \\
r1-distill-7b-gsm8k & 0.621 & 0.574 & \textbf{0.369} & 0.412 \\
r1-distill-7b-math500 & 0.764 & 0.704 & \textbf{0.518} & 0.521 \\
\bottomrule
\end{tabular}
\end{table}

The truncated area narrows the result without reversing it. Final-step entropy beats the binary flag in six cells, the exceptions being qwen7b-gsm8k and llama31-8b-gsm8k, the two most accurate cells in the panel. The violation count beats the flag in six, ties it to three decimals on mistral7b-gsm8k, and trails it by 0.001 on qwen7b-gsm8k, where the flag is the best of the four. Under either range, final-step entropy beats the violation count in six cells, with the same two GSM8K cells as the exceptions. So the original's report that monotonicity dominates all cheap scalar baselines does not reproduce on this panel under either range. It holds on qwen7b-gsm8k under both, which is the cell whose conditions match the original's, and on llama31-8b-gsm8k under truncation only. Scalar coherence behaves as the original describes only on GSM8K, where under either range it is the worst of the four in all three standard cells; on MATH-500 it beats the binary flag everywhere, which is the same setting-dependence Section~\ref{sec:magnitude} reports for the H2 criterion.

\subsection{Logistic Regression}

A logistic regression over the standardized monotonicity flag, trajectory length and final entropy tells the same story about which term carries the weight. The coefficient on final entropy runs from $-0.55$ to $-1.43$ and is the largest in every cell; the coefficient on monotonicity runs from $-0.11$ to $+0.47$; the coefficient on trajectory length stays between $-0.41$ and $+0.28$. The original's own regression on its pilot gives $+0.36$, $-0.05$ and $-0.07$ respectively, so final entropy is nearly inert there and dominant here.

\section{The Trajectory-Length Artifact}
\label{app:artifact}

Trajectory length in the stored records counts prefixes that yielded an extractable answer, not segmentation units, so it cannot by itself locate the step cap. At the 600-token budget 21.0\% of anchor MATH-500 chains reach the maximum trajectory length of 9 points, and at 1280 only 3.2\% do, which was first read as the cap relaxing. It cannot be. A stopping cap can only increase unit counts, so a genuine fall would indicate something wrong in the generation path.

The scans supply the correct reading and a traced run confirms it directly. The ninth trajectory point appears almost exclusively where the reference chain was truncated mid-reasoning, with odds ratios of 12.2 at the 600-token budget and 19.7 at 1280, because a truncated chain leaves the continuation something to finish while a complete chain ends in its own answer and the continuation adds nothing parseable. In the traced run of Appendix~\ref{app:traced} the association is close to deterministic. A trajectory reaches the maximum of nine points on 52\% of truncated chains and 3\% of complete ones. That run also shows the mechanism is not what the phrase extraction loss suggests. Of 71 prefixes across 741 that yielded no answer, 70 returned an empty string and one returned text the extractor could not parse, because a chain ending in a boxed answer leaves the continuation nothing to say. The spike at eight points is what complete chains do, and the higher nine-count at 600 was the artifact.

\section{Execution and Provenance}
\label{app:execution}

The eight confirmatory cells ran on a rented A40 between 2026-08-30 and 2026-09-01, at a measured steady-state throughput of 21.7 seconds per problem, so that a 1319-problem cell takes roughly eight and a half hours. Recorded wall clocks include 8h32m for the anchor on GSM8K, 5h45m for the anchor on MATH-500, 6h34m for Mistral on GSM8K, and 9.7h and 5.6h for the two Llama-3.1 cells. Every run writes a provenance manifest recording the harness commit, the resolved model commit SHA, the GPU, the CUDA and torch versions and the package set; in each of the eight cells the resolved model commit matches the pinned revision exactly, which confirms independently that the pin describes the weights actually served.

The manifests record the working tree as dirty for every cell. The salvaged pod contents show four untracked analysis scripts at the repository root, the earliest created before the first cell began, which accounts for a non-empty \texttt{git status --porcelain} without any tracked source file differing. The flag itself cannot distinguish the two cases, so this account rests on the salvaged file timestamps rather than on the flag.

\section{The Monotonicity Tolerance}
\label{app:tolerance}

The tolerance $\epsilon$ was registered at 0.01 with a pre-specified sweep over 0, 0.05 and 0.10. Recomputing every cell at each value moves the accuracy gap by at most 1.88 points anywhere in the panel, and $\epsilon$=0 and $\epsilon$=0.01 give results identical to the last digit in all eight cells. This reproduces the original's own ablation, where the monotone rate is 0.737 at every $\epsilon$ at or below 0.10.

The mechanism appears to be the estimator's discreteness rather than the criterion's robustness. With m=5 samples the answer entropy can only take values realizable by a partition of five or fewer parsed answers, and across the eight cells it takes 13 distinct values with a minimum spacing of 0.0152 nats. Of the 9,487 step-to-step entropy rises observed in the panel, 181, or 1.9\%, are at or below 0.10 nats. Any tolerance below 0.0152 therefore reclassifies nothing at all, and any tolerance up to 0.10 can reach at most 1.9\% of transitions. The original attributes the same stability to entropy jumps tending to exceed 0.20 nats, which is consistent with this measurement and less specific about why.

The practical reading is that the $\epsilon$ sweep is not an informative robustness check at m=5, since its outcome is close to arithmetically determined, and that a sweep over m would test the same concern more usefully. Running the m sweep the registration already pre-specifies is recommended before the tolerance result is cited as evidence of robustness.

\section{Analyses Not Performed}
\label{app:notrun}

Six pre-specified secondary and exploratory analyses were not run.

Segmentation-rule sensitivity across newline, sentence and 40-token-window rules is the most consequential omission, and the traced run makes its likely result predictable rather than unknown, since every excluded chain in that sample segments under the newline rule. Running it on the registered cells would convert that from strongly supported to measured. Three-seed robustness of the primary cells, the $m$ sweep over 10 and 20, and the quantization comparison of 4-bit against fp16 on the anchor all require fresh generation. The replication of the original's log-probability calibration finding and the logit-based entropy variant are blocked differently. The harness never recorded token log-probabilities, so both need a code change as well as a rerun. All six remain runnable, and the segmentation sweep and the $m$ sweep are the two recommended first.

\section{Traced Run}
\label{app:traced}

None of the analyses above can show what the protocol does to an individual problem, because the panel stored counts and derived statistics and no text. One hundred problems were therefore rerun on each of two cells, the anchor on MATH-500 (qwen7b-math500) and Mistral on GSM8K (mistral7b-gsm8k), with the reference chain, the segmentation units and every sampled continuation retained. The configuration is otherwise the registered one for each cell, the problems are indices 0 to 99 of each benchmark, fixed in advance, and the run is illustrative. It supports no registered hypothesis and enters no tally.

It supplies two measurements the rest of the study could not. The first is the composition of the lost trajectory points, reported in Section~\ref{sec:segmentation}. The second is the cause of the segmentation exclusions, also in Section~\ref{sec:segmentation}, which turns out to be a disagreement between the prompt and the segmenter rather than a property of any model. Neither changes a registered result.

\end{document}